\documentclass[lettersize,journal]{IEEEtran}
\usepackage{amsmath,amsfonts}
\usepackage{algorithmic}
\usepackage{algorithm}
\usepackage{array}
\usepackage[caption=false,font=normalsize,labelfont=sf,textfont=sf]{subfig}
\usepackage{textcomp}
\usepackage{stfloats}
\usepackage{url}
\usepackage{verbatim}
\usepackage{graphicx}
\usepackage{cite}
\usepackage{hhline}
\usepackage{makecell}
\usepackage{booktabs,multirow}
\usepackage{hyperref}
\usepackage[table]{xcolor}
\begin{document}

\title{PSTTS: A Plug-and-Play Token Selector for Efficient Event-based Spatio-temporal Representation Learning}


\author{{Xiangmo~Zhao,~Nan~Yang,~Yang~Wang,~Zhanwen~Liu}}

\markboth{Journal of \LaTeX\ Class Files,~Vol.~14, No.~8, August~2021}%
{Shell \MakeLowercase{\textit{et al.}}: A Sample Article Using IEEEtran.cls for IEEE Journals}


\maketitle



\begin{abstract}
Mainstream event-based spatio-temporal representation learning methods typically process event streams by converting them into sequences of event frames, achieving remarkable performance. However, they neglect the high spatial sparsity and inter-frame motion redundancy inherent in event frame sequences, leading to significant computational overhead. Existing token sparsification methods for RGB videos rely on unreliable intermediate token representations and neglect the influence of event noise, making them ineffective for direct application to event data. In this paper, we propose Progressive Spatio-Temporal Token Selection (PSTTS), a Plug-and-Play module for event data without introducing any additional parameters. PSTTS exploits the spatio-temporal distribution characteristics embedded in raw event data to effectively identify and discard spatio-temporal redundant tokens, achieving an optimal trade-off between accuracy and efficiency. Specifically, PSTTS consists of two stages: Spatial Token Purification and Temporal Token Selection. Spatial Token Purification discards noise and non-event regions by assessing the spatio-temporal consistency of events within each event frame to prevent interference with subsequent temporal redundancy evaluation. Temporal Token Selection evaluates the motion pattern similarity between adjacent event frames, precisely identifying and removing redundant temporal information. We apply PSTTS to four representative backbones UniformerV2, VideoSwin, EVMamba, and ExACT on the HARDVS, DailyDVS-200, and SeACT datasets. Experimental results demonstrate that PSTTS achieves significant efficiency improvements. Specifically, PSTTS reduces FLOPs by 29–43.6\% and increases FPS by 21.6–41.3\% on the DailyDVS-200 dataset, while maintaining task accuracy. Our code will be available.

\end{abstract}


\section{Introduction}

\IEEEPARstart{I}{n} recent years, event cameras have attracted considerable attention due to their notable advantages, including high dynamic range, high temporal resolution, and low power consumption \cite{gallego2020event,finateu20205,brandli2014240,liu2024enhancing}. Unlike frame-based conventional cameras, event cameras asynchronously capture brightness changes, effectively filtering out redundant static background information and providing richer temporal cues for moving objects. Moreover, event cameras primarily capture objects' edges, which alleviates privacy concerns related to attributes such as skin color and gender \cite{qian2025ucf}. Benefiting from these advantages, event cameras have demonstrated exceptional performance in spatio-temporal vision tasks such as action recognition \cite{wang2024dailydvs,wang2024hardvs,zhou2024exact,wan2022s2n} and video anomaly detection \cite{qian2025ucf}, and hold promise as an ideal solution for power-constrained scenarios, such as surveillance.

This has inspired a wide range of spatio-temporal representation learning methods for event data. These methods can be broadly categorized into asynchronous and synchronous approaches based on how they process event streams. (1) Asynchronous methods \cite{zhou2022spikformer,yao2023spike} exploit the asynchronous nature of event streams, applying bio-inspired spiking neural networks (SNNs) to utilize the high temporal resolution of event data. However, due to the immaturity of these models, they have yielded suboptimal performance and exhibited limited adaptability. (2) Synchronous approaches \cite{wang2024hardvs,wan2022s2n,peng2023get,qian2025ucf,zhou2024exact,wang2024event} typically segment the event stream into fixed time intervals and convert it into image-like tensor sequences, enabling the spatio-temporal representation modeling with mature deep neural networks such as CNN, ViT, and SSM. Although ViT-based and SSM-based methods have demonstrated promising performance, there exists an inherent trade-off between computational efficiency and the temporal resolution of the input sequence. To maintain reasonable computational cost, existing methods are forced to use relatively large time windows (\emph{e.g.}, 0.5 seconds) for event stream sampling, as well as significantly downsample the input sequence to a low frame rate (\emph{e.g.}, eight frames), which compromises the high temporal resolution of the event stream and limits their performance \cite{wang2024dailydvs}.


\begin{figure}[tbp]
\centering
\includegraphics[scale=0.33]{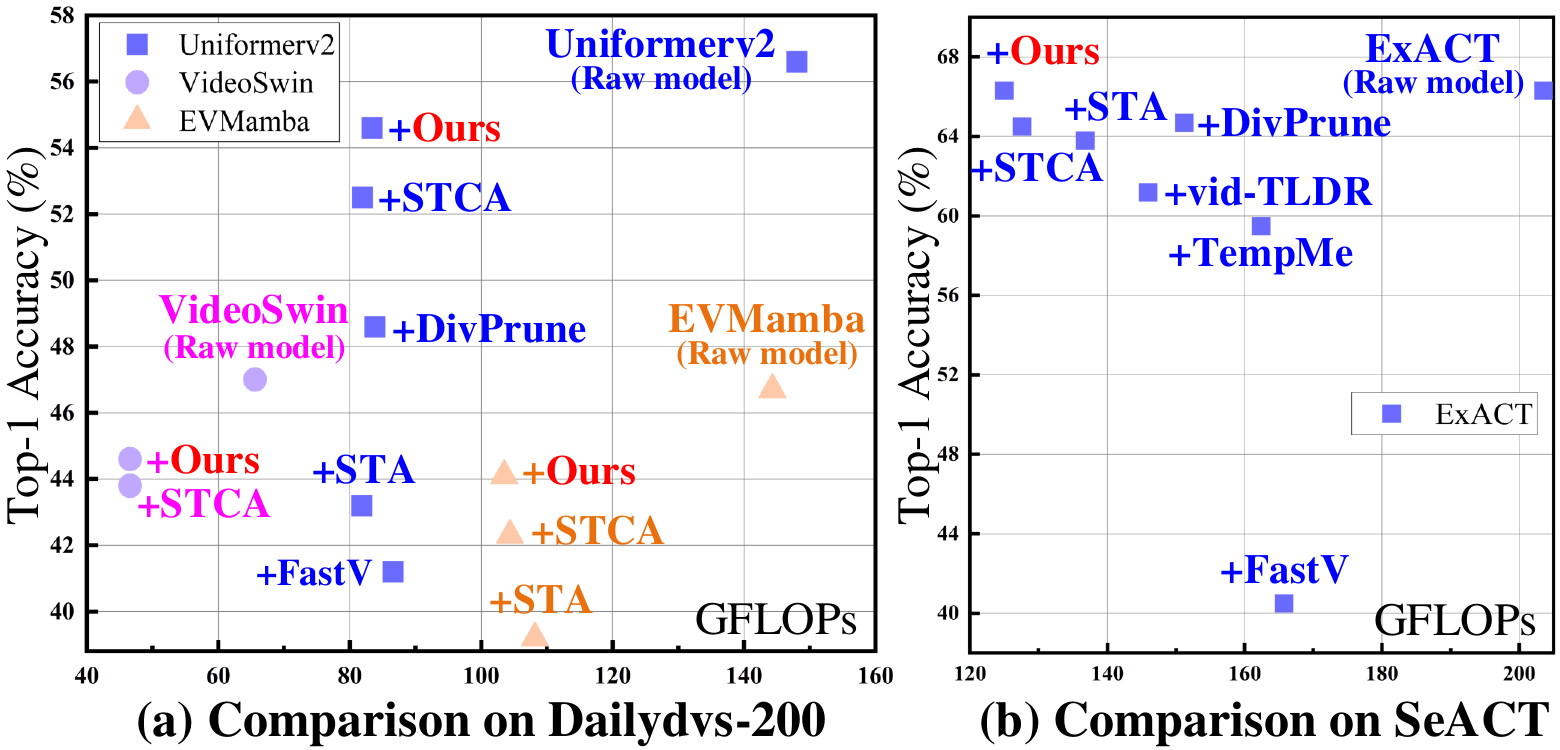}
\caption{(a) and (b): Comparison of accuracy and FLOPs after applying state-of-the-art token sparsification methods and our PSTTS to four different architectures (UniformerV2, VideoSwin, EVMamba, and ExACT) on the Dailydvs-200 and ExACT datasets. Our method achieves a superior balance between accuracy and efficiency.}
\label{fig1}
\end{figure}

\begin{figure}[tbp]
\centering
\includegraphics[scale=0.32]{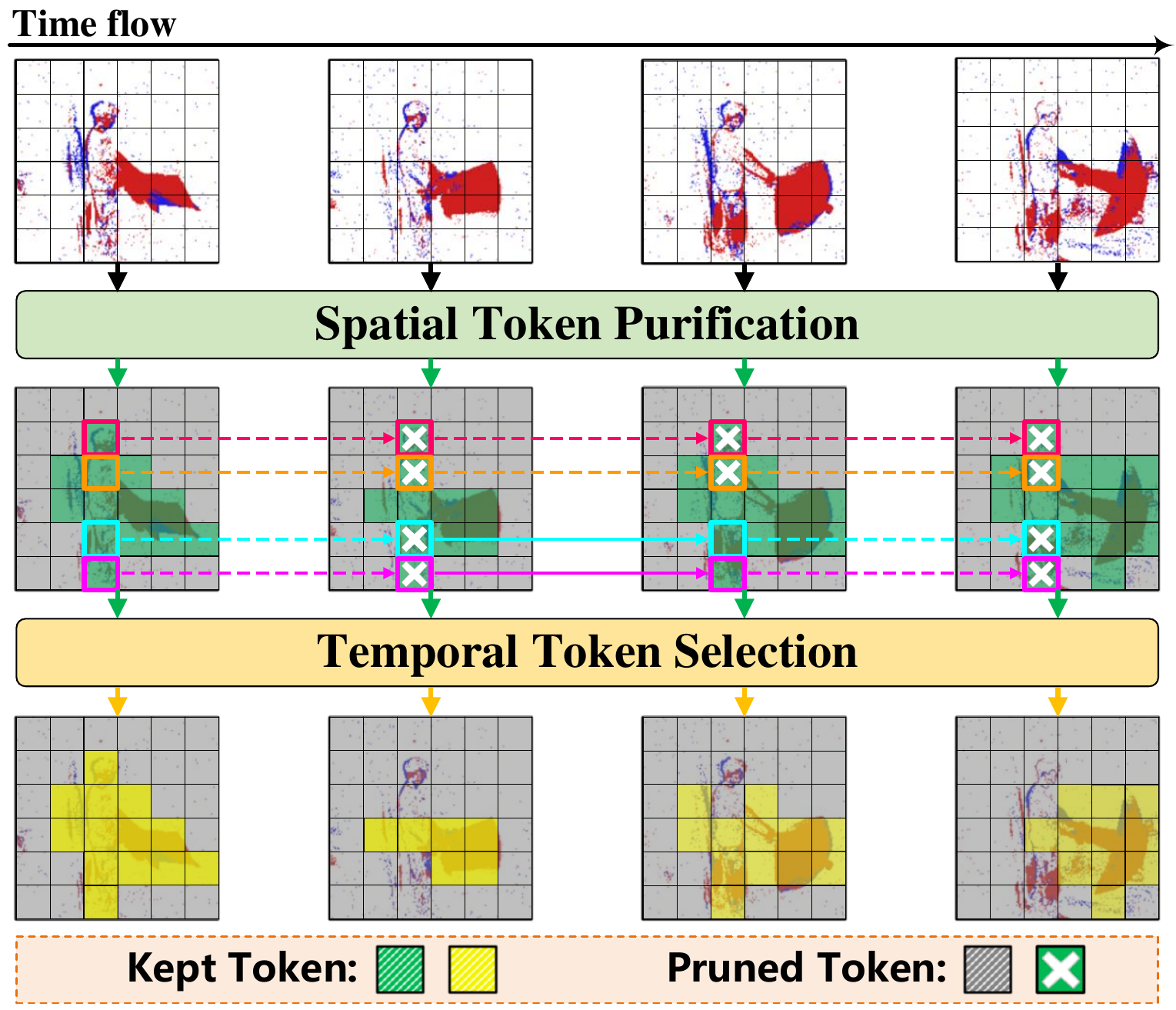}
\caption{A conceptual overview of our PSTTS, which consists of Spatial Token Purification and Temporal Token Selection. Spatial Token Purification first identifies and removes non-event and noise regions. Subsequently, Temporal Token Selection evaluates the inter-frame motion similarity of the remaining active event regions (green areas), eliminates redundant motion patterns (\emph{e.g.}, head and torso movement), and retains important spatiotemporal information (yellow areas).}
\label{fig11}
\end{figure}




One promising solution to this problem is to reduce the number of input tokens. Event data have been proven to exhibit high sparsity in spatial dimension \cite{peng2024scene,yang2025smamba} and high redundancy of inter-frame motion in temporal dimension \cite{zhou2024exact}. However, there is no effective method to address such spatio-temporal redundancy for event data. It is worth noting that in the RGB modality, researchers have already proposed various token sparsification methods \cite{wang2022efficient,ding2023prune,choi2024vid,chen2024image,shen2024tempme,alvar2025divprune,li2024vidtome} for video data to reduce computational costs by pruning spatio-temporal redundant tokens. However, directly applying these methods to event data presents two critical challenges: (1) Existing approaches typically assess token importance by extracting intermediate token embeddings from model layers or by designing learnable scoring modules. However, their scoring methods face two key limitations. First, its accuracy heavily depends on the discriminability of token representations, which is closely related to the quality of the event frames \cite{liu2024enhancing}. When the sampling interval is too large or the object motion is too intense, the event frames may exhibit blurred structural information, reducing the discriminability of the token representations and the accuracy of the scoring results. Second, token representations in the early and middle layers are insufficiently encoded, failing to accurately reflect tokens' actual contribution to the task \cite{wu2020visual,chefer2021transformer,alvar2025divprune}. (2) In addition, event data contain a large amount of background noise caused by immature circuit designs \cite{ding2023mlb,wan2022s2n,duan2024led}, which can interfere with the evaluation of token importance and lead to imprecise pruning results.

In this paper, we propose a Plug-and-Play module Progressive Spatio-Temporal Token Selection (PSTTS) designed for event data, as shown in Fig.\ref{fig11}. PSTTS does not rely on unreliable intermediate feature representations within the model. Instead, it effectively identifies and discards spatio-temporal redundant tokens by exploiting the spatio-temporal consistency of active events and the spatio-temporal distribution similarity between adjacent event frames, which are inherent in the raw event data, thereby improving computational efficiency, as shown in Fig.\ref{fig1}(a, b). Specifically, PSTTS consists of two stages: Spatial Token Purification and Temporal Token Selection. In the Spatial Token Purification stage, the spatio-temporal consistency of the event stream within each frame sampling interval is assessed to identify spatiotemporally inconsistent noise and non-event regions, eliminating their influence on subsequent temporal redundancy estimation. In the Temporal Token Selection stage, the redundant motion patterns are identified and discarded by evaluating the motion magnitude similarity and trajectory shape similarity across different regions of adjacent event frames, further reducing computational overhead. To illustrate the effectiveness of our method, we conduct extensive experiments on two large-scale event-based action recognition datasets HARDVS \cite{wang2024hardvs} and Dailydvs-200 \cite{wang2024dailydvs}, and one event-language based action recognition dataset SeACT \cite{zhou2024exact} using UniformerV2 \cite{li2023uniformerv2}, VideoSwin \cite{liu2022video}, EVMamba \cite{wang2024event} and ExACT \cite{zhou2024exact} as backbones. 

Overall, our contribution can be summarized as follows:

(1) We propose PSTTS, the first spatio-temporal token sparsification method specifically designed for event data. PSTTS progressively identifies and discards spatially and temporally redundant tokens, effectively reducing the computational burden of spatio-temporal representation modeling. 

(2) We design two token sparsification stages: Spatial Token Purification and Temporal Token Selection. The first stage removes noise and non-event regions by assessing the spatio-temporal consistency of events. The second stage eliminates redundant motion patterns by evaluating the spatio-temporal distribution similarity between adjacent event frames.


(3) The experimental results demonstrate that PSTTS can effectively improve efficiency while maintaining accuracy and can be applied to different model architectures, surpassing current state-of-the-art methods and exhibiting excellent generalization capability.


\section{Related Work}

This section provides an overview of event-based spatio-temporal representation learning methods, followed by a review of recent advancements in token sparsification methods.

\subsection{Spatio-temporal Representation Learning for Event Stream}
In spatio-temporal vision tasks, event-based spatio-temporal representation learning methods can be categorized into asynchronous and synchronous approaches based on the input event representations. Asynchronous methods \cite{amir2017low,george2020reservoir,liu2021event,zhou2022spikformer,yao2023spike} process event data as spike streams and utilize spiking neural networks (SNNs) to model the spatio-temporal characteristics of event streams, which align well with their inherent asynchronous nature. However, due to the immaturity of current model architectures, these methods yield suboptimal performance. Synchronous methods convert event streams into continuous, image-like frames and adapt the processing paradigms and mature architectures of the RGB modality, including CNN-based \cite{gao2023action,pradhan2019n,wan2022s2n}, ViT-based \cite{sabater2022event,xie2024event,peng2023get,zhou2024exact}, and SSM-based \cite{wang2024event} models. Among these, ViT-based and SSM-based methods achieve superior performance due to their larger receptive fields. However, they overlook the high spatial sparsity of event data and the temporal redundancy caused by repetitive motion across frames, resulting in computational redundancy. To address this issue, ExACT \cite{zhou2024exact} proposes an Adaptive Fine-grained Event Representation to sample keyframes from redundant frame sequences, thereby improving computational efficiency. However, it still requires downsampling of the generated keyframes to avoid excessive computational overhead.

In this paper, we propose a Plug-and-Play token sparsification method PSTTS that reduces the computational cost of ViT-based and SSM-based methods by pruning tokens with high spatio-temporal redundancy. This approach can be integrated with keyframe sampling techniques to further improve efficiency.

\subsection{Token sparsification}
In image classification and object detection tasks, several methods have been proposed to improve computational efficiency by removing redundant tokens for image \cite{rao2021dynamicvit,liu2022adaptive,chen2023diffrate,chen2023sparsevit,liu2024revisiting} or event data \cite{peng2024scene,yang2025smamba}. Although these frame-level approaches can be directly applied to each frame of a video to support spatio-temporal vision tasks, they fail to consider temporal dependencies and redundancy across frames, which may disrupt the spatio-temporal structure of the video and result in limited sparsification \cite{wang2022efficient}. 

Recently, some methods specifically designed for RGB videos have been proposed to incorporate temporal information into the token sparsification process. Token merge methods \cite{choi2024vid, shen2024tempme, li2024vidtome} propose to merge redundant spatio-temporal tokens based on their similarity. However, they fail to restore the merged tokens to their original spatial dimensions directly, limiting their applicability to hierarchical models and restricting their generalization capability. In token pruning methods, they typically assess token importance based on token embeddings. STTS \cite{wang2022efficient} designs a scoring network to evaluate token importance and removes redundant tokens using a spatiotemporally decoupled strategy. STA \cite{ding2023prune} introduces a semantic-aware temporal accumulation scoring method that assesses spatio-temporal redundancy based on key values in the self-attention mechanism. FastV \cite{chen2024image} identifies and prunes unimportant tokens based on their average attention scores, thereby improving inference efficiency in VLM (Vision-Language Model) tasks. DivPrune \cite{alvar2025divprune} retains a maximally diverse subset of tokens by computing the cosine similarity between token embeddings. However, spatio-temporal token sparsification for event data remains an unexplored area. Directly applying RGB-domain methods to event data results in significant performance degradation, primarily due to the unreliable token representation in early and intermediate layers. Furthermore, the high noise levels inherent in event streams significantly interfere with assessing token importance.

\section{Methods}
\begin{figure*}[tbp]
\centering
\includegraphics[scale=0.75]{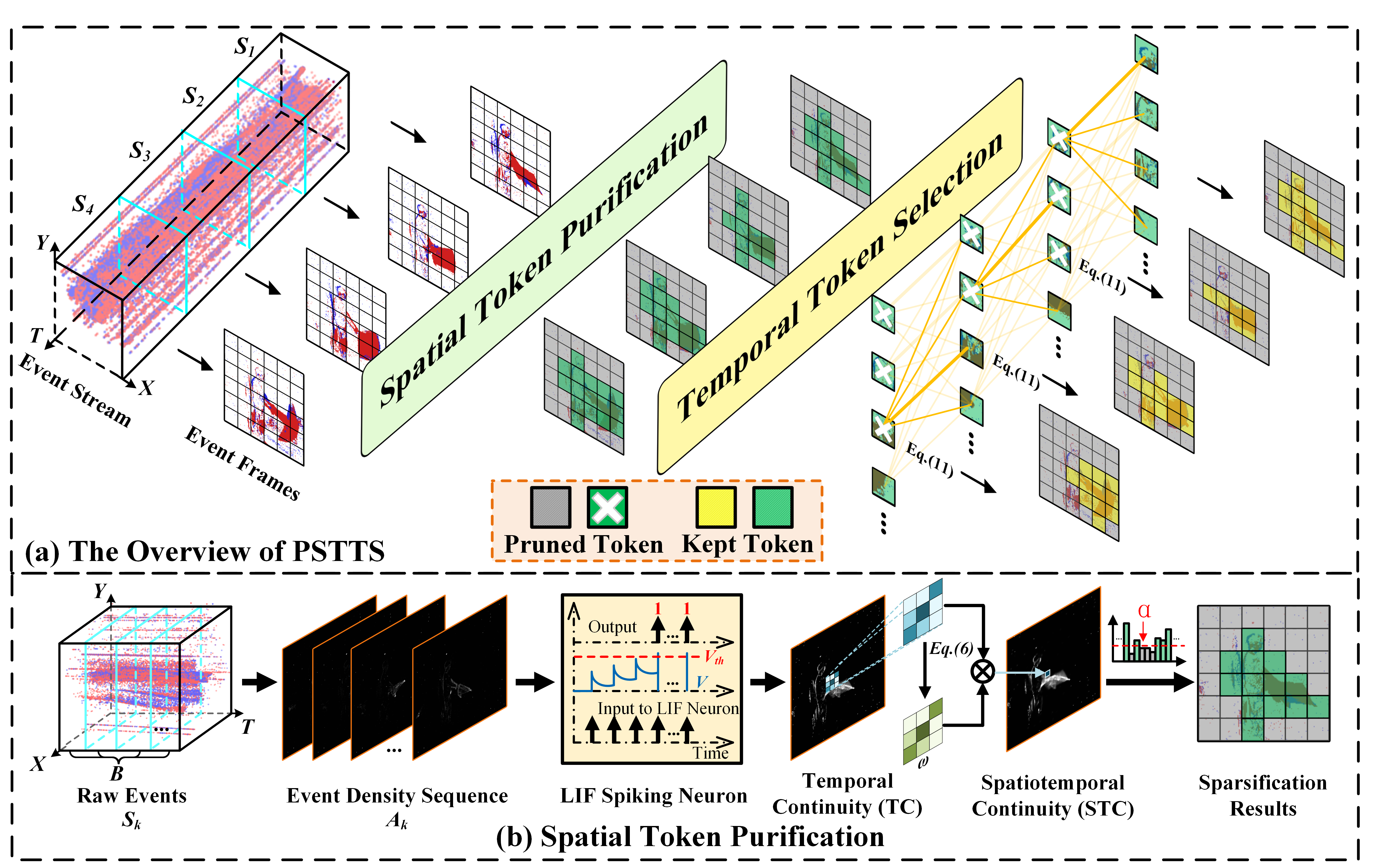}
\caption{\textbf{The framework of our PSTTS.} The input event stream is first divided into segments at a fixed time interval to form an event-frame sequence. Meanwhile, the raw event segments are then processed by the Spatial Token Purification module independently, which identifies and removes noise and non‑event regions in each frame through spatio-temporal consistency analysis. The remaining tokens (green regions) are subsequently passed to the Temporal Token Selection module to assess temporal redundancy and discard repeated motion patterns across frames. Finally, the retained tokens (yellow regions) are fed into the model for spatio-temporal representation learning.}
\label{fig2}
\end{figure*}

Event frame sequences contain extensive noise and non-event regions in the spatial dimension, as well as repetitive inter-frame motion patterns in the temporal dimension, resulting in significant computational redundancy during spatio-temporal representation learning. To mitigate this issue, we propose a Plug-and-Play module Progressive Spatio-Temporal Token Selection (PSTTS), which consists of two stages: Spatial Token Purification and Temporal Token Selection, as shown in Fig.\ref{fig2}. Firstly, in the spatial dimension, Spatial Token Purification independently identifies and discards interference from noise and non-event regions within each frame, thereby improving the quality of subsequent processing. Secondly, in the temporal dimension, the Temporal Token Selection assesses the motion magnitude and trajectory shape similarity of active event regions across adjacent frames to identify repetitive motion patterns, further reducing redundant computation.


\subsection{Spatial Token Purification}
The inherent circuitry characteristics of event cameras generate substantial event noise, which is often temporally discontinuous or spatially isolated \cite{guo2022low,yang2025smamba}. In contrast, active events are typically triggered along the edges of objects due to motion or illumination changes and exhibit spatio-temporal consistency. Based on this spatio-temporal distribution difference, we design the Spatial Token Purification (STP) stage that accurately identifies active event regions by modeling temporal dynamics and spatial continuity, effectively eliminating the interference of noise and non-event regions in the subsequent assessment of temporal redundancy, as shown in Fig.\ref{fig2}(b).

Given the event stream $S$, we follow \cite{wang2024event,wang2024dailydvs,zhou2024exact} and divide it into $K$ segments $S=\{S_1,...,S_K\}$ at a fixed time interval. Each event segment $S_k$ contains the event data $\left\{\left(x_{i}, y_{i}, t_{i}, p_{i}\right)\right\}_{i=1}^{N}$, where $(x_i,y_i)$ is the spatial coordinates, $t_i$ represents the timestamp, $p_i\in\{{-1,1}\}$ indicates the event polarity, $N$ is the number of events. After that, the spatio-temporal consistency of each event segment $S_k$ is independently evaluated. 

Firstly, the temporal dynamic of events triggered at the same pixel is analyzed to identify temporally discontinuous noise. Due to the microsecond-level temporal resolution of the asynchronous event stream, directly processing raw events leads to excessive computational complexity. Therefore, each event segment is further divided into $B$ discrete temporal windows at a smaller fixed interval, and the number of events in each pixel within each window is then accumulated to form an event density sequence ${A_{k}}\in{\mathbb{R}^{B \times H \times W}}$, enabling a coarse-grained representation that balances evaluation precision and computational efficiency. The expression is as follows:
\begin{equation} 
{A_{k}[b]}\left(x,y\right)=\sum_{e_{i} \in \mathcal{E}}\delta\left(x-x_{i},y-y_{i}\right),
\end{equation}
where $\mathcal{E}$ denotes the event set within the $b$-th temporal bin, and $\delta$ represents the Kronecker delta.

Subsequently, the window-level event density sequence $A_{k}$ is fed into a Leaky Integrate-and-Fire (LIF) spiking neuron, which directly encodes the temporal relationships between windows and achieves accurate capture of continuity and causal dependencies in the event stream. Additionally, the temporal activation frequency of hot pixel noise is constrained to the window level, thereby suppressing the inference from such noise with high triggering frequency. The formulation is as follows:
\begin{equation} 
V_k[b]=V_k[b-1]+\frac{1}{\tau}\left(A_{k}[b]-\left(V_k[b-1]-V_\mathrm{reset}\right)\right),
\end{equation}
\begin{equation} 
Spike_k[b]=\Theta\left(V_k[b]-V_{\mathrm{th}}\right),\quad V_k[b]\leftarrow V_\mathrm{reset},
\end{equation}
where $\tau$ is the membrane time constant. $\Theta(\cdot)$ denotes the Heaviside step function that fires a spike (outputs 1) if $V_k[b]-V_{\mathrm{th}}\geq0$, and $V_k[b]$ is the membrane potential which resets to $V_\mathrm{reset}$ if a spike is fired. The accumulated temporal activations of the LIF spiking neuron serve as the score of temporal continuity, as described below:
\begin{equation}
TC_{k} = \sum_{b = 1}^{B} Spike_{k}[b]
\end{equation}

Secondly, spatial continuity is further analyzed by aggregating local neighborhood information to identify spatially isolated noise. In the neighborhood of active events, spatially adjacent events are often triggered by the same moving edge and tend to exhibit similar temporal continuity due to the coherence of object motion. Based on this observation, we construct joint spatial-temporal weights to assign greater importance to neighboring events that are both spatially close and temporally consistent, thereby enhancing motion-related spatio-temporal patterns. The process is as follows:
\begin{equation}
STC_{k}\left(x_{i}, y_{i}\right) = \frac{\sum_{j \in \mathcal{N}_{i}} w_{ij} \cdot TC_{k}\left(x_{j}, y_{j}\right)}{\sum_{j \in \mathcal{N}_{i}} w_{ij}},
\end{equation}
\begin{equation}
\begin{split}
w_{ij} = \exp\biggl(
& -\frac{\left\|\left(x_{i}, y_{i}\right)-\left(x_{j}, y_{j}\right)\right\|^{2}}{2\sigma_{s}^{2}} \\
& -\frac{\left|TC_{k}\left(x_{i}, y_{i}\right)-TC_{k}\left(x_{j}, y_{j}\right)\right|^{2}}{2\sigma_{t}^{2}}
\biggr),
\end{split}
\end{equation}
where $\mathcal{N}_{i}$ is the neighborhood of $\left(x_{i}, y_{i}\right)$, $\sigma_{s}^{2}$ and $\sigma_{t}^{2}$ represent the spatial variance and temporal variance, respectively. Average pooling is then applied to downsample $STC_k$ by a factor of $p$, where $p$ denotes the patch size used during tokenization. The resulting $STC_k^{down}\in{\mathbb{R}^{H/p \times W/p}}$ indicates the spatio-temporal continuity of events within the region corresponding to each token.


Finally, in the token selection stage, the average of the spatio-temporal continuity map $STC_k^{down}$ is employed as the sparsification threshold $\alpha$:

\begin{equation}
{\alpha} = \frac {sum \left(STC_k^{down}\right)} {\frac{HW}{p^{2}}}.
\end{equation}

Regions with scores below the threshold are identified as non-event or noise regions and are subsequently removed. This approach enables adaptive adjustment of the threshold and the number of retained tokens based on the complexity of each sample, thereby providing a certain degree of sample-level adaptivity. Meanwhile, the spatio-temporal continuity map captures the spatio-temporal distribution characteristics of different regions and reveals their motion patterns, thereby facilitating the sparsification process in the second stage.


\subsection{Temporal Token Selection}
Repetitive motion regions often appear in event frame sequences, introducing redundant computations in spatio-temporal modeling. To address this issue, we design the Temporal Token Selection (TTS) stage, which evaluates the motion magnitude and trajectory shape similarity of different regions in adjacent frames based on the spatio-temporal continuity map, identifying redundant motion patterns.


Firstly, the spatio-temporal continuity score maps $STC$ are divided into patches of size $p \times p$. Then, based on the sparsification results in the first stage, the retained patches $P_k\in{\mathbb{R}^{M \times p \times p}}$ of $STC$ are extracted for motion pattern similarity calculation, eliminating the influence of noise and non-event regions, $M$ represents the number of retained patches. 

\begin{table*}[tbp]
\small
\centering
\renewcommand\arraystretch{1.2} 
\caption{Comparisons with the SOTA methods on the HARDVS dataset. * denotes the results after fine-tuning.}
\begin{tabular}{lcccccc}
     
    \Xhline{1.5pt}
    \textbf{Methods} & \textbf{Pub.} & \textbf{Backbone} & \textbf{top-1(\%)} & \textbf{top-5(\%)} & \textbf{GFLOPs$\times$views}  & \textbf{FPS} \\
    \hline
    SlowFast\cite{feichtenhofer2019slowfast} & ICCV'19 & CNN & 46.5 & 54.8 & 6.95$\times$3$\times$4 & 19.1 \\
    TSM\cite{lin2019tsm} & ICCV'19 & CNN & 52.6 & 60.1 & 33$\times$3$\times$1 & 50.5 \\
    ESTF\cite{wang2024hardvs} & AAAI'24 & CNN & 51.2 & 57.5 & - & - \\
    TimeSformer\cite{bertasius2021space} & ICML'21 & ViT & 50.8 & 58.7 & 141$\times$3$\times$1 & 16.7 \\
    \hline
    \rowcolor{gray!20}VideoSwin\cite{liu2022video} & CVPR'22 & SwinT & 53.3 & 63.1 & 65.6$\times$3$\times$1 & 30.1 \\
    +STCA\cite{yang2025smamba}  & AAAI'25 & - & 51.3(-2.0) & 61.0(-2.1) & 46.7$\times$3$\times$1(-28.8\%) & 35.0(+16.3\%) \\
    \rowcolor{orange!20}\textbf{+Ours} & - & - & \textbf{52.5}(-0.8) & \textbf{62.1}(-1.0) & \textbf{46.7}$\times$3$\times$1(-28.8\%) & \textbf{36.3}(+20.6\%) \\
    \hline
    \rowcolor{gray!20}EVMamba\cite{wang2024event} & ICLR'25 & SSM & 54.1 & 63.3 & 144.3 & 14.8 \\
    +STA*\cite{ding2023prune} & ICCV'23 & - & 54.0(-0.1) & \textbf{64.0}(+0.7) & 108.2(-25.0\%) & 9.5(-35.8\%) \\
    +STCA*\cite{yang2025smamba} & AAAI'25 & - & 53.9(-0.2) & 63.0(-0.3) & \textbf{105.5}(-26.9\%) & 16.2(+9.5\%) \\
    \rowcolor{orange!20}\textbf{+Ours*} & - & - & \textbf{54.4}(+0.3) & \textbf{64.0}(+0.7) & 108.7(-24.7\%) & \textbf{16.8}(+13.5\%) \\
    \hline
    \rowcolor{gray!20}Uniformerv2\cite{li2023uniformerv2} & ICCV'23 & ViT & 54.6 & 65.2 & 148 & 41.8 \\
    +STA\cite{ding2023prune} & ICCV'23 & - & 52.2(-2.4) & 62.9(-2.3) & 85.7(-42.1\%) & 43.5(+4.1\%) \\
    +FastV\cite{chen2024image} & ECCV'24 & - & 52.7(-1.9) & 64.2(-1.0) & 86.6(-41.5\%) & \textbf{64.2}(+53.6\%) \\
    +DivPrune\cite{alvar2025divprune} & CVPR'25 & - & 47.5(-7.1) & 59.7(-5.5) & 86.1(-41.8\%) & 45.3(+8.4\%) \\
    +STCA\cite{yang2025smamba} & AAAI'25 & - & 52.0(-2.6) & 62.7(-2.5) & 85.9(-42.0\%) & 58.0(+38.8\%) \\
    \rowcolor{orange!20} \textbf{Ours} & - & - & \textbf{53.7}(-0.9) & \textbf{64.6}(-0.6) & \textbf{85.6}(-42.2\%) & 56.8(+35.9\%) \\

    \Xhline{1.5pt}
\end{tabular}
\label{table1}
\end{table*} 

\textbf{Motion Magnitude Similarity.} Similar motion patterns tend to trigger a comparable number of events and exhibit similar spatio-temporal continuity within spatio-temporal neighborhoods, reflecting similar motion magnitudes. To this end, we design a motion magnitude similarity metric which measures the correlation between the average motion magnitude of corresponding patches across adjacent frames:
\begin{equation} 
MMS_{k}^{i} = \sum_{j=1}^{M} \frac{2\mu_{k}^{i} \cdot \mu_{k-1}^{j}}{(\mu_{k}^{i})^{2} + (\mu_{k-1}^{j})^{2}},
\end{equation}
where $\mu_{k}^{i}$ denotes the mean value of the $i$-th patch in $P_k$. A higher value of $MMS_{k}^{i}$ approaching 1 indicates greater similarity in motion magnitude between the two patches.

\textbf{Trajectory Shape Similarity.} Similar motion patterns tend to exhibit similar trajectory shape. To this end, we design a trajectory shape similarity metric, which measures the structural correlation of spatio-temporal distribution between retained patches in adjacent frames, defined as follows:
\begin{equation}
TSS_{k}^{i} = \sum_{j=1}^{M} \frac{\delta_{k,k-1}^{i,j}}{\delta_{k}^{i} \cdot \delta_{k-1}^{j}},
\end{equation}
where $\delta_{k,k-1}^{i,j}$ denotes the covariance between the two patches, and $\delta_{k}^{i}$ represents the standard deviation.

Finally, the temporal motion redundancy score is defined as follows:
\begin{equation}
TMR_{k}^{i} =(1 - MMS_{k}^{i} \cdot TSS_{k}^{i})
\end{equation}

In the second stage, to preserve regions with redundant motion patterns that are still important for the task, the L2 norm of token embeddings is introduced as an auxiliary metric to assess semantic importance. Additionally, the spatio-temporal continuity score is preserved to suppress potential interference from residual noise remaining from the first stage. The token importance score $Score$ in the second stage is defined as follows:
\begin{equation}
Score_{k}^{i} = STC_{k}^{\text{down},i} \cdot TMR_{k}^{i} \cdot L2_{k}^{i},
\end{equation}
where $L2_{k}^{i}$ is L2 norm of $i$-th retained token in the $k$-th frame. As in the first stage, the average of the $Score$ is used as the threshold to remove temporally redundant regions.

\section{EXPERIMENTS}
\subsection{Experimental Setup}

\textbf{Datasets.} We evaluate our method on three event-based action recognition datasets: 1) \textbf{HARDVS} \cite{wang2024hardvs} is currently the largest dataset for event-based action recognition, comprising 107,646 recordings of 300 action categories. 2) \textbf{DailyDVS-200} \cite{wang2024dailydvs} comprises 22,046 recordings across 200 action categories, with action durations ranging from 1 to 20 seconds and annotations for 14 attributes. 3) \textbf{SeACT} \cite{zhou2024exact} provides detailed caption-level labels of each action for event-text action recognition and contains 58 actions under four themes.

\textbf{Implementation details.} 
To demonstrate that our PSTTS is generic for different architectures, we implement it using four model architectures with state-of-the-art (SOTA) performance, including the ViT-based UniformerV2 \cite{li2023uniformerv2}, the window-based Transformer VideoSwin \cite{liu2022video}, the SSM-based EVMamba \cite{wang2024event}, and the vision-language model ExACT \cite{zhou2024exact}. For UniformerV2, VideoSwin, and ExACT, we use their pretrained weights and simply apply our method directly after the intermediate layers, evaluating them without any additional training. For EVMamba, directly testing with the token sparsification method causes severe performance degradation due to SSM's sequence-sensitive characteristics \cite{zhan2024exploring, wu2025dynamic}. Therefore, we finetune the pretrained EVMamba with our method. We measure the computational cost via FLOPs (floating-point operations) and FPS (clips/s) at a batch size of 1 on a single NVIDIA-3090. For different datasets, we adopt their original event frame generation methods to ensure a fair comparison. Specifically, we sample event frames at 0.25 seconds and 0.5 seconds intervals on the HARDVS and DailyDVS-200 datasets, respectively, and set the number of input event frames to 8. For the SeACT dataset, we adopt the adaptive frame sampling strategy proposed by ExACT to generate the input event frame sequences.



\begin{table*}[tbp]
\small
\centering
\renewcommand\arraystretch{1.2} 
\caption{Comparisons with the SOTA methods on the DailyDVS-200 dataset. * denotes the results after fine-tuning.}
\begin{tabular}{lcccccc}
     
    \Xhline{1.5pt}
    \textbf{Methods} & \textbf{Pub.} & \textbf{Backbone} & \textbf{top-1(\%)} & \textbf{top-5(\%)} & \textbf{GFLOPs$\times$views}  & \textbf{FPS} \\
    \hline
    SlowFast\cite{feichtenhofer2019slowfast} & ICCV'19 & CNN & 41.5 & 68.2 & 6.95$\times$3$\times$4 & 19.1 \\
    TSM\cite{lin2019tsm} & ICCV'19 & CNN & 40.9 & 71.5 & 33$\times$3$\times$1 & 50.5 \\
    ESTF\cite{wang2024hardvs} & AAAI'24 & CNN & 24.7 & 50.2 & - & - \\
    TimeSformer\cite{bertasius2021space} & ICML'21 & ViT & 44.3 & 74.0 & 141$\times$3$\times$1 & 16.7 \\
    GET\cite{peng2023get} & ICCV'23 & SwinT & 37.3 & 61.6 & - & - \\
    Spikformer\cite{zhou2022spikformer} & ICLR'23 & SNN, ViT & 36.9 & 62.4 & - & - \\
    \hline
    \rowcolor{gray!20}VideoSwin\cite{liu2022video} & CVPR'22 & SwinT & 47.0 & 73.6 & 65.6$\times$3$\times$1 & 30.1 \\
    +STCA\cite{yang2025smamba} & AAAI'25 & - & 43.8(-3.2) & 70.6(-3.0) & \textbf{46.6}$\times$3$\times$1(-29.0\%) & 35.6(+18.3\%) \\
    \rowcolor{orange!20}\textbf{+Ours} & - & - & \textbf{44.6}(-2.4) & \textbf{71.5}(-2.1) & \textbf{46.6}$\times$3$\times$1(-29.0\%) & \textbf{36.6}(+21.6\%) \\
    \hline
    \rowcolor{gray!20}EVMamba\cite{wang2024event} & ICLR'25 & SSM & 46.7 & 73.0 & 144.3 & 14.8 \\
    +STA*\cite{ding2023prune} & ICCV'23 & - & 39.2(-7.5) & 66.3(-6.7) & 108.2(-25.0\%) & 9.5(-35.8\%) \\
    +STCA*\cite{yang2025smamba} & AAAI'25 & - & 42.3(-4.4) & 71.0(-2.0) & 104.4(-27.7\%) & 18.9(+27.7\%) \\
    \rowcolor{orange!20}\textbf{+Ours*} & - & - & \textbf{44.1}(-2.6) & \textbf{72.3}(-0.7) & \textbf{103.5}(-28.3\%) & \textbf{19.2}(+29.7\%) \\
    \hline
    \rowcolor{gray!20}Uniformerv2\cite{li2023uniformerv2} & ICCV'23 & ViT & 56.6 & 81.5 & 148 & 41.6 \\
    +STA\cite{ding2023prune} & ICCV'23 & - & 43.2(-13.4) & 67.4(-14.1) & \textbf{81.8}(-44.7\%) & 44.2(+6.3\%) \\
    +FastV\cite{chen2024image} & ECCV'24 & - & 41.2(-15.4) & 66.2(-15.3) & 86.6(-41.5\%) & \textbf{64.2}(+54.3\%) \\
    +DivPrune\cite{alvar2025divprune} & CVPR'25 & - & 48.6(-8.0) & 72.1(-9.4) &  83.8(-43.4\%) & 47.1(+13.2\%) \\
    +STCA\cite{yang2025smamba} & AAAI'25 & - & 52.5(-4.1) & 77.1(-3.8) & 81.9(-44.7\%) & 59.0(+41.8\%) \\
    \rowcolor{orange!20}\textbf{+Ours} & - & - & \textbf{54.6}(-2.0) & \textbf{79.3}(-2.3) & 83.4(-43.6\%) & 58.8(+41.3\%) \\

    \Xhline{1.5pt}
\end{tabular}
\label{table2}
\end{table*}

     

\begin{table}[tbp]
\small
\centering
\renewcommand\arraystretch{1.2} 
\caption{Comparisons with the SOTA methods on the SeAct dataset.}
\scalebox{1.0}{
\begin{tabular}{lccccc}
     
    \Xhline{1.5pt}
    \textbf{Methods} & \textbf{Pub.} & \textbf{top-1(\%)} & \textbf{GFLOPs} \\
    \hline
    EventTransAct\cite{de2023eventtransact} & IROS'23 & 57.8 & -  \\
    EvT\cite{sabater2022event} & CVPR'22 & 61.3 & - \\
    \hline
    \rowcolor{gray!20}ExACT\cite{zhou2024exact} & CVPR'24 & 66.3  & 203.5 \\
    +STA\cite{ding2023prune} & ICCV'23 & 63.8(-2.5) & 136.8(-32.8\%) \\
    +FastV\cite{chen2024image} & ECCV'24 & 40.5(-25.8) & 165.7(-18.6\%)  \\
    +DivPrune\cite{alvar2025divprune} & CVPR'25 & 64.7(-1.6) & 151.2(-25.7\%) \\
    +vid-TLDR\cite{choi2024vid} & CVPR'24 & 61.2(-5.1) & 145.9(-28.3\%) \\
    +TempMe\cite{shen2024tempme} & ICLR'25 & 59.5(-6.8) & 162.4(-20.2\%) \\
    +STCA\cite{yang2025smamba} & AAAI'25 & 64.5(-1.8) & 127.6(-37.3\%) \\
    \rowcolor{orange!20}+Ours & -& \textbf{66.3}  & \textbf{125.0}(-38.6\%)\\
    \Xhline{1.5pt}
\end{tabular}} 
\label{table3}
\end{table}

\subsection{Main Results}
To evaluate the superiority of our method, we compare our PSTTS with SOTA token sparsification methods, including RGB-domain, event-domain, and VLM-domain methods in three datasets. Furthermore, to demonstrate the effectiveness under higher input frame rates, we conduct additional comparative experiments on the Dailydvs-200 dataset which contains long event recordings.

\textbf{Comparison with SOTA methods.} Firstly, we compare our method with four token pruning methods, including three RGB-domain methods (STA \cite{ding2023prune}, FastV \cite{chen2024image}, and DivPrune \cite{alvar2025divprune}) and one event-domain method (STCA \cite{yang2025smamba}) using the VideoSwin, EVMamba, and UniformerV2 backbones on the HARDVS and DailyDVS-200 datasets. Due to the constraints of the window attention operation in VideoSwin, only the STCA method is compared for VideoSwin. For a fair comparison, we maintain similar FLOPs for all methods. 

The results on the HARDVS dataset are shown in Tab.\ref{table1}, directly applying our method achieves FLOPs reductions of 28.8\% for VideoSwin and 42.2\% for UniformerV2, along with speed improvements of 20.6\% and 35.9\%, respectively, at the cost of less than a 1\% drop in accuracy, significantly outperforming other token pruning methods. Furthermore, fine‑tuning EVMamba with our method yields an additional improvement of 0.3\% in top‑1 accuracy and 0.7\% in top‑5 accuracy, while reducing FLOPs by 24.7\% and achieving a 13.5\% speedup. 

The results on the DailyDVS‑200 dataset are presented in Tab.\ref{table2}, our method exhibits even greater advantages under a larger event‑frame sampling interval (0.5 s). For instance, when applied to UniformerV2, the top-1 accuracy decreases by only 2.0\%, in contrast to the STA and FastV methods, which suffer significant drops of 13.4\% and 15.4\%, respectively. This is because STA and FastV identify redundant tokens based on similarity or attention scores between token embeddings. However, the event frames in the DailyDVS-200 dataset have a large sampling interval of 0.5 seconds, resulting in blurred structures and poor semantic distinctiveness of token embeddings, which in turn leads to inaccurate sparsification results. In contrast, our method leverages the spatio-temporal statistical characteristics of event data to effectively identify redundant spatio-temporal information, thereby resisting interference from low-quality input data and adapting to different datasets and model architectures.

Secondly, we compare our method with the four aforementioned token pruning methods as well as two token merging methods, vid-TLDR \cite{choi2024vid} and TempMe \cite{shen2024tempme} using the ExACT backbone on the SeACT dataset. As shown in Tab.\ref{table3}, although ExACT significantly reduces temporal redundancy through adaptive keyframe sampling, incorporating our method still achieves a 38.6\% reduction in FLOPs without any loss of accuracy, significantly outperforming other methods.

Notably, STCA, which also leverages the statistical characteristics of event data, achieves suboptimal performance, further validating the soundness of our chosen direction. Moreover, our method enables more accurate preservation of critical information by precisely assessing the spatio-temporal continuity and temporal redundancy of event streams. Overall, our method achieves superior performance across various model architectures and datasets while effectively reducing computational redundancy.




\begin{table*}[tbp]
\small
\centering
\renewcommand\arraystretch{1.2} 
\caption{Comparisons with the SOTA methods on the DailyDVS-200 dataset with higher frame rate input. * denotes the results after fine-tuning.}
\begin{tabular}{lcccccc}
     
    \Xhline{1.5pt}
    \textbf{Methods} & \textbf{Pub.} & \textbf{Backbone} & \textbf{top-1(\%)} & \textbf{top-5(\%)} & \textbf{GFLOPs$\times$views}  & \textbf{FPS} \\
    \hline
    \rowcolor{gray!20}VideoSwin & CVPR'22 & SwinT & 55.6 & 79.8 & 141$\times$3$\times$1 & 20.6 \\
    +STCA & AAAI'25 & - & 52.4(-3.2) & 77.3(-2.5) & \textbf{100}$\times$3$\times$1(-29.1\%) & 24.0(+16.5\%) \\
    \rowcolor{orange!20}\textbf{+Ours} & - & - & \textbf{53.7}(-1.9) & \textbf{78.3}(-1.5) &  \textbf{100}$\times$3$\times$1(-29.1\%) & \textbf{24.6}(+19.4\%) \\
    \hline
    \rowcolor{gray!20}EVMamba & ICLR'25 & SSM & 51.7 & 78.1 & 251.3 & 7.5 \\
    +STA*\cite{ding2023prune} & ICCV'23 & - & 46.4(-5.3) & 73.2(-4.9) & 199(-20.8\%) & 5.6(-25.3\%) \\
    +STCA*\cite{yang2025smamba} & AAAI'25 & - & 48.9(-2.8) & 76.0(-2.1) & 196(-22.0\%) & 9.5(+26.7\%) \\
    \rowcolor{orange!20}\textbf{+Ours*} & - & - & \textbf{50.1}(-1.6) & \textbf{76.8}(-1.3) & \textbf{196}(-22.0\%) & \textbf{10.0}(+33.3\%) \\
    \hline
    \rowcolor{gray!20}Uniformerv2 & ICCV'23 & ViT & 63.5 & 86.1 & 297 & 21.8 \\
    +STA & ICCV'23 & - & 53.5(-10.0) & 77.8(-8.3) & 167(-43.8\%) & 24.8(+13.8\%) \\
    +FastV & ECCV'24 & - & 45.3(-18.2) & 71.3(-14.8) & 167(-43.8\%) & \textbf{34.5}(+58.3\%) \\
    +DivPrune & CVPR'25 & - & 57.9(-5.6) & 80.5(-5.6) &  166(-44.1\%) & 30.0(+37.6\%) \\
    +STCA & AAAI'25 & - & 60.1(-3.4) & 82.7(-3.4) & \textbf{165}(-44.4\%) & 33.0(+51.4\%) \\
    \rowcolor{orange!20}\textbf{+Ours} & - & - & \textbf{61.6}(-1.9) & \textbf{84.1}(-2.0) & \textbf{165}(-44.4\%) & 32.0(+46.8\%) \\

    \Xhline{1.5pt}
\end{tabular}
\label{table4}
\end{table*}

\begin{table}[tbp]
\centering
\small
\renewcommand\arraystretch{1.1} 
  \caption{THE PERFORMANCE OF different scoring metrics.}
  \begin{tabular} {cc|ccc|cc}
    \Xhline{1.5pt}
     \multicolumn{2}{c|}{ \textbf{STP}} & \multicolumn{3}{c|}{ \textbf{TTS}}& {\textbf{HARDVS}} & {\textbf{Dailydvs-200}} \\ 
     \textbf{TC} & \textbf{STC} &\textbf{TMR} & \textbf{STC} & \textbf{L2} & top1/top5 & top1/top5\\
    \hline
    {\checkmark} & &{\checkmark} & {\checkmark} & {\checkmark} & 52.0/62.1 & 53.1/78.4   \\
    {\checkmark} &{\checkmark} &{\checkmark} & && 53.2/\textbf{64.3} & 54.1/78.7   \\
    {\checkmark} &{\checkmark} &{\checkmark} & {\checkmark} & &  53.4/64.2 & 54.3/79.0   \\
     {\checkmark} &{\checkmark} &{\checkmark} & {\checkmark} & {\checkmark} &  \textbf{53.7/64.3} & \textbf{54.6/79.3}  \\
    \Xhline{1.5pt}
  \end{tabular}
  \label{table5}
\end{table}

\begin{figure}[tbp]
\centering
\includegraphics[scale=0.53]{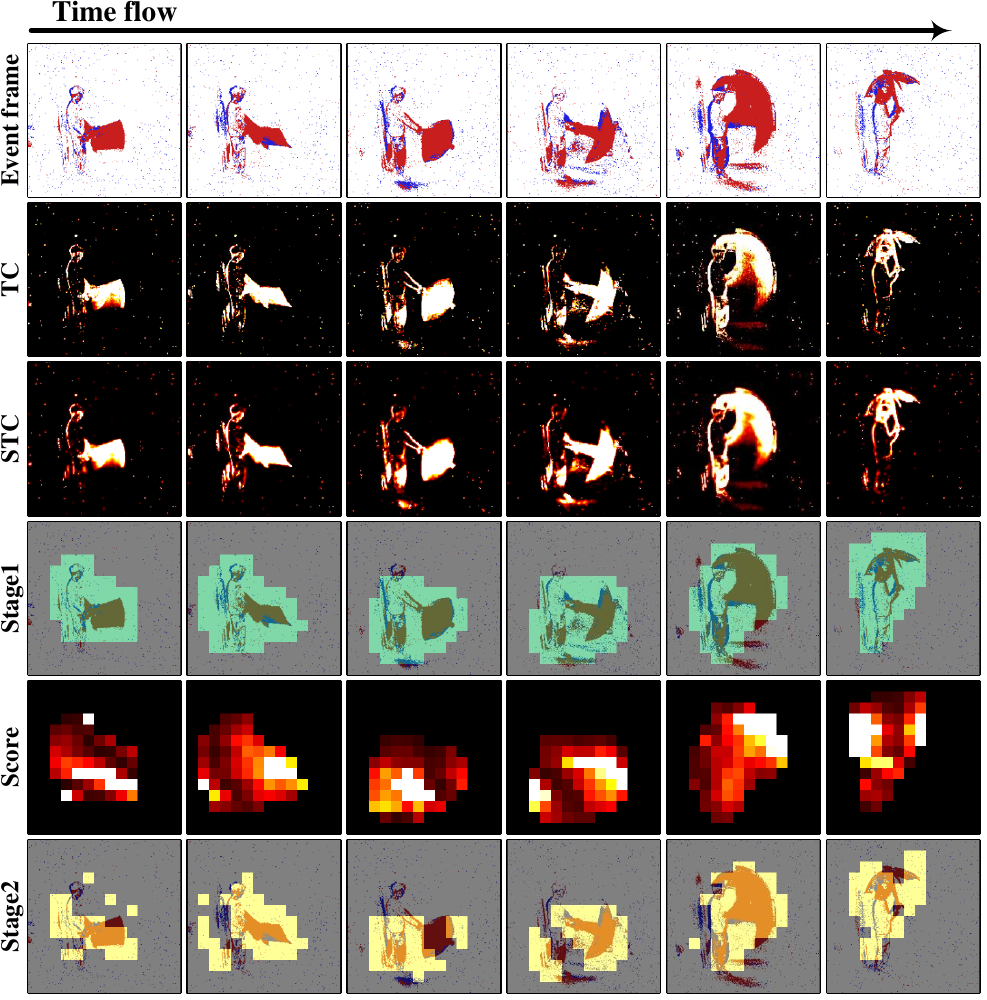}
\caption{Visualizations of raw event frames of action ``open umbrella'', temporal continuity map, spatio-temporal continuity map, sparsification results of stage1, score map and sparsification results of stage2.}
\label{fig3}
\end{figure}

\textbf{Results with higher frame rate input.} To evaluate the performance of our method under higher input frame rates, we conduct experiments on the DailyDVS-200 dataset, which contains sufficiently long event streams. Specifically, we sample the event sequences from DailyDVS-200 at a shorter interval of 0.25 seconds to generate higher frame-rate input sequences and adjust the number of event frames fed into the model to 16. As shown in Tab.\ref{table4}, when the input event sequence is longer, the baseline models achieve much better performance, for instance, from 56.6 to 63.5 for Uniformerv2. Meanwhile, under similar FLOPs reduction ratio, the performance gap between our method and the original performance is also reduced. This indicates that with higher input frame rates, the benefits of our method are amplified, presenting a promising approach for efficiently processing long-range event streams at scale.


\subsection{Ablation Study}
In this section, we conduct an ablation study to analyze the impact of various core components of our method based on UniformerV2 on the HARDVS and Dailydvs-200 datasets.

\textbf{Contribution of different scoring metrics.} To explore the contributions of different scoring metrics in the two stages, we conduct experiments under similar FLOPs for various scoring strategies. The results are presented in Tab.\ref{table5}, comparing the first row and last row demonstrates that in the first stage, the joint spatial-temporal weights effectively measure the spatial continuity of events and identify spatially independent noise. Further, incorporating temporal motion redundancy, spatio-temporal continuity, and semantics in the second stage helps to discover informative tokens.



\textbf{Contribution of the progressive sparsification.} To explore the contributions of the two proposed token sparsification stages, we conduct experiments under similar FLOPs for different sparsification stages. As shown in Tab.\ref{table6}, applying only the Spatial Token Purification fails to remove temporally redundant motion patterns and may incorrectly discard critical but event-sparse regions under high sparsity scenes. Applying only the Temporal Token Selection is susceptible to interference from noise and non-event regions, failing to accurately assess the similarity of motion patterns across frames.

The score maps and sparsification results of different stages are shown in Fig.\ref{fig3}. In row 2, TC effectively measures the temporal continuity of the event stream and suppresses temporally discontinuous noise. In row 3, STC further effectively identifies and suppresses spatially independent noise. In row 4, the Spatial Token Purification successfully identifies active event regions while removing non-event and noisy regions. In row 5, the Temporal Token Selection successfully detects redundant motion patterns across frames, further reducing the regions requiring computation.


\begin{table}[tbp]
\centering
\small
\renewcommand\arraystretch{1.1} 
  \caption{THE Contribution of two-stage sparsification.}
  \begin{tabular} {c|cc}
    \Xhline{1.5pt}
    {\multirow{2}*{\textbf{Method}}} & \textbf{HARDVS} & \textbf{Dailydvs-200} \\ 
    & top1/top5 & top1/top5\\
    \hline
    with STP  & 53.0/64.0 & 53.5/78.8   \\
     with TTS &  53.1/63.8 & 53.4/78.4  \\
     with all & \textbf{53.7/64.3} & \textbf{54.6/79.3}    \\
    \Xhline{1.5pt}
  \end{tabular}
  \label{table6}
\end{table}

\textbf{Token selection strategy.} We conduct an ablation study to analyze the impact of different token selection strategies on performance, including the commonly used fixed sparsity ratio strategy and our proposed adaptive sparsity threshold strategy. For the fixed sparsity ratio strategy, we first calculate the average token discard ratios of PSTTS and then apply these ratios to remove the same number of tokens for all samples. As shown in Tab.\ref{table7}, the adaptive sparsity threshold strategy can select an appropriate number of tokens to discard for each sample, avoiding under-sparsification for simple samples or over-sparsification for complex ones, demonstrating superior performance. Furthermore, even when adopting the same fixed sparsity ratio strategy as the SOTA methods, our method still achieves better performance, benefiting from the precise identification of spatio-temporal redundancy by PSTTS.

\subsection{Sparsification Visualizations}
\begin{figure}[tbp]
\centering
\includegraphics[scale=0.51 ]{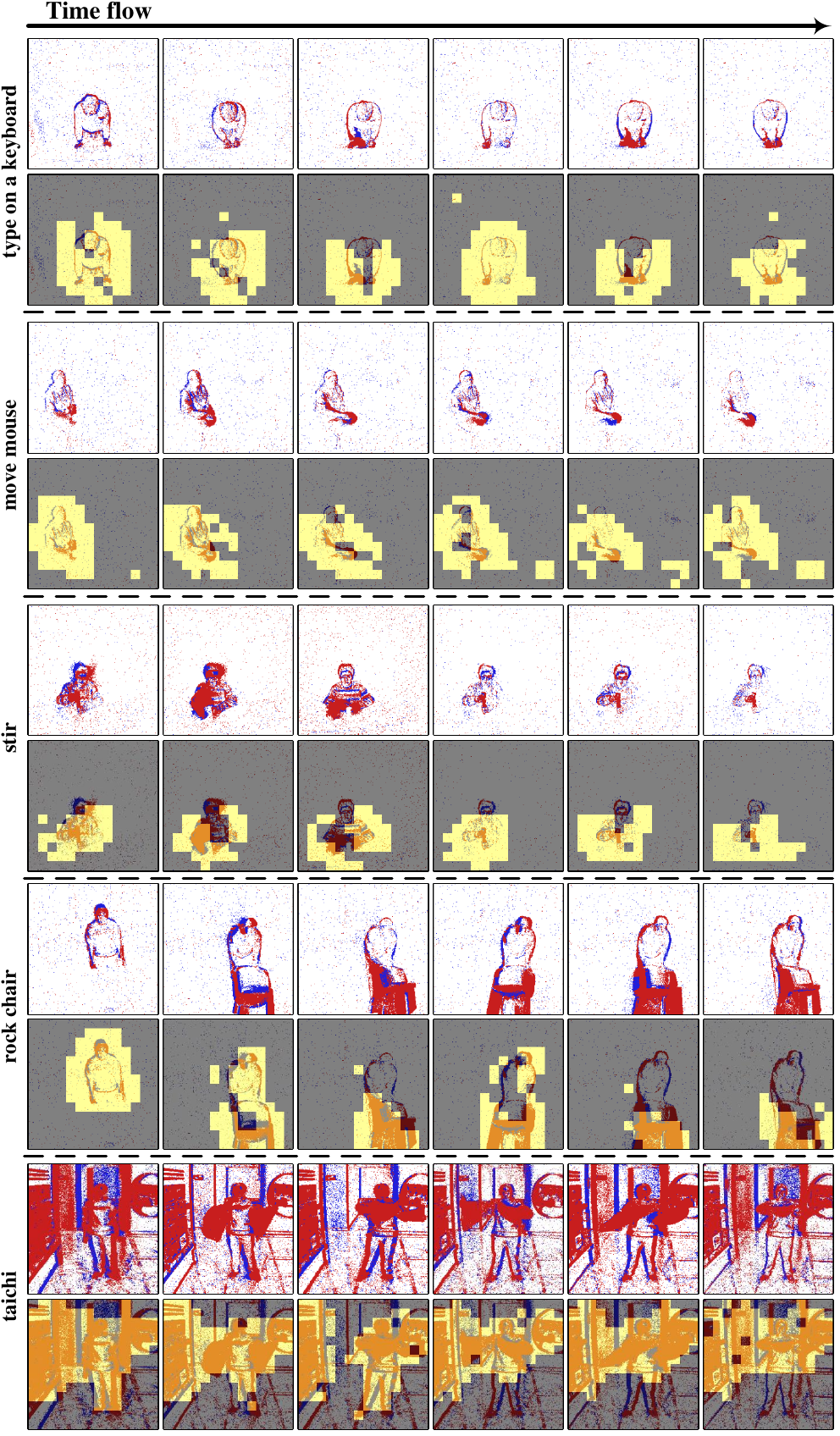}
\caption{Visualizations of raw event frames and kept tokens of action ``type on a keyboard'', ``move mouse'', ``stir'', ``rock chair'' and ``taichi''.}
\label{fig4}
\end{figure}

The sparsification results of ``type on a keyboard'', ``move mouse'', ``stir'', ``rock chair'' and ``taichi'' are shown in Fig.\ref{fig4}. The action groups represent small-scale motion, large-scale motion, and intense background motion, with increasing levels of scene complexity. Our method effectively preserves key informative regions while removing the spatio-temporal redundant information, exhibiting robust scene-adaptive capability.




\begin{table}[tbp]
\centering
\small
\renewcommand\arraystretch{1.1} 
  \caption{THE PERFORMANCE of Different token selection strategies.}
  \begin{tabular} {c|cc}
    \Xhline{1.5pt}
    {\multirow{2}*{\textbf{Method}}} & \textbf{HARDVS} & \textbf{Dailydvs-200} \\ 
    & top1/top5 & top1/top5\\
    \hline
    Fixed sparsity ratio  & 53.4/64.0 & 53.9/78.7   \\
    Adaptive sparsity threshold & \textbf{53.7/64.3} & \textbf{54.6/79.3}    \\
    \Xhline{1.5pt}
  \end{tabular}
  \label{table7}
\end{table}

\section{Conclusion}
In this paper, we propose PSTTS, the first token sparsification method tailored for event-based spatio-temporal vision tasks. PSTTS progressively filters out noise and non-event regions, and discards temporally redundant motion patterns, significantly improving the efficiency of spatio-temporal representation modeling. Experiments on the HARDVS, DailyDVS-200, and SeACT datasets demonstrate that PSTTS can be seamlessly integrated into different model architectures (\emph{e.g.}, UniformerV2, VideoSwin, EVMamba, and ExACT) and achieves superior performance compared to SOTA token sparsification methods. Meanwhile, PSTTS supports finer-grained partitioning of the event stream and can process longer event frame sequences with richer temporal information at relatively low cost, enabling the more effective utilization of the high temporal resolution of event data, achieving an optimal balance between accuracy and efficiency. 



\bibliographystyle{IEEEtran}
\bibliography{trans}

\end{document}